%% file: main.tex
\documentclass[letterpaper, 10 pt, conference]{ieeeconf}  

\IEEEoverridecommandlockouts                              

\usepackage{graphicx}
\usepackage{array}
\usepackage{booktabs}
\usepackage{graphicx, amsmath}
\usepackage{adjustbox}
\usepackage{caption}
\usepackage{multirow}
\usepackage{subcaption}
\usepackage[table,xcdraw]{xcolor}
\usepackage{subcaption} 
\usepackage{times}
\usepackage{latexsym}
\usepackage{xspace}
\usepackage{amsmath}
\usepackage{amsmath} 
\usepackage{amssymb}  
\usepackage{mathtools}
\usepackage{mathrsfs}
\usepackage{pifont}
\DeclareMathOperator*{\argmin}{arg\,min}

\title{\LARGE \bf
Safety Aware Task Planning via Large Language Models in Robotics}

\author{Azal Ahmad Khan$^{1*}$, Michael Andrev$^{1*}$, Muhammad Ali Murtaza$^{2}$, Sergio Aguilera$^{3}$,\\
Rui Zhang$^{1}$, Jie Ding$^{1}$, Seth Hutchinson$^{2}$, Ali Anwar$^{1}$
\thanks{*Equal contributions}
\thanks{$^{1}$University of Minnesota -- Twin Cities {\tt\small Email: \{khan1086, zhan@1386, dingj, aanwar\}@umn.edu}}%
\thanks{$^{2}$ Institute for Robotics and Intelligent Machines (IRIM), Georgia Institute of Technology, Atlanta, GA, USA {\tt\small Email: \{mamurtaza, seth\}@gatech.edu}}
\thanks{$^{3}$ Department of Mechanical and Metallurgical Engineering, School of Engineering, Pontificia Universidad Cat\'olica de Chile, Santiago, Chile{\tt\small Email:  sfaguile@uc.cl} }
}

\begin{document}

\maketitle
\thispagestyle{empty}
\pagestyle{empty}

\newcommand{\proj}{SAFER\xspace}

\begin{abstract}
The integration of large language models (LLMs) into robotic task planning has unlocked better reasoning capabilities for complex, long-horizon workflows. However, ensuring safety in LLM-driven plans remains a critical challenge, as these models often prioritize task completion over risk mitigation. This paper introduces \proj (\underline{S}afety-\underline{A}ware \underline{F}ramework for \underline{E}xecution in \underline{R}obotics), a multi-LLM framework designed to embed safety awareness into robotic task planning. SAFER employs a Safety Agent that operates alongside the primary task planner, providing safety feedback.  Additionally, we introduce LLM-as-a-Judge, a novel metric leveraging LLMs as evaluators to quantify safety violations within generated task plans. Our framework integrates safety feedback at multiple stages of execution, enabling real-time risk assessment, proactive error correction, and transparent safety evaluation. We also integrate a control framework using Control Barrier Functions (CBFs) to ensure safety guarantees within SAFER’s task planning. We evaluated \proj against state-of-the-art LLM planners on complex long-horizon tasks involving heterogeneous robotic agents, demonstrating its effectiveness in reducing safety violations while maintaining task efficiency. We also verify the task planner and safety planner through actual hardware experiments involving multiple robots and a human.
\end{abstract}

\input{Sections/Introduction}
\input{Sections/Related_Works}
\input{Sections/Framework}
\input{Sections/controlbarrierfunction}
\input{Sections/Experiments}
\input{Sections/DiscussionConclusion}










\bibliographystyle{ieeetr}
\bibliography{references}

\end{document}

%% file: Sections/Introduction.tex
\section{Introduction}
\label{sec:Introduction}

Large Language Models (LLMs) have become pivotal in the advancement of robotic task planning, allowing systems to interpret unstructured instructions, reason about multiagent collaborations, and generate context-aware action sequences~\cite{liu2024coherent, zhang2023building, duan2024aha, dogan2024grace, roychoudhuryefficient, sikorski2024deployment, latif2024physicsassistant}. Recent frameworks exemplify this progress, where LLMs orchestrate heterogeneous robots~\cite{liu2024coherent, kannan2023smart, sun2024interactive, mu2024embodiedgpt}. By leveraging LLM's commonsense reasoning and ability to parse natural language, LLMs bridge the gap between high-level human commands and low-level robotic execution, outperforming traditional logic-based planners~\cite{chen2025scar}. These capabilities are particularly transformative for long-horizon tasks that require dynamic coordination between robots with distinct action spaces, such as transporting objects across cluttered environments~\cite{zeng2023large}.

\textbf{Safety Challenges in LLM-based Task Planning. }Despite these advances, safety remains a significant challenge in LLM-based systems~\cite{wu2024highlighting, qi2024safety}. Traditional planners rely on rigid constraints and predefined rules to mitigate risks, but LLMs prioritize task efficiency, often overlooking hazards like spatial conflicts, partial action sequences, or unstable object handoffs~\cite{deng2025proactive}. For instance, an LLM might direct a quadrotor to land on a checkout counter while a robotic arm is still active, risking collisions, or generate a plan where a robotic arm executes a [grab] action but omits the critical [place] step. These issues are worsen in long-horizon tasks, where the planner’s limited context window forces a trade-off between retaining historical steps for coherence and incorporating safety guidelines. Without explicit safeguards, LLMs inherently lack the prioritization of risk mitigation, leaving systems vulnerable to catastrophic failures.

\textbf{Multi-LLM Conversation in Planning Module. }To address this, we propose \proj a \underline{S}afety-\underline{A}ware \underline{F}ramework for \underline{E}xecution in \underline{R}obotics.
\proj utilizes multi-LLM collaboration for safety-aware planning. Instead of burdening a single LLM with both task planning and safety checks, we decouple these roles. As shown in the Fig.~\ref{fig:framework} our framework introduces a Safety Planning LLM that operates alongside the central task planner by providing feedback and enforcing generalized safety rules, such as avoidance of conflict in the workspace and validation of the action sequence, without inflating the context window of the task planner. This LLM intervenes during the planning loop, by providing feedback for unsafe actions (e.g., adding prerequisite checks or retiming steps) while preserving task goals. Furthermore, we formalize LLM-as-a-Judge~\cite{gu2024survey}, a safety metric where a dedicated LLM evaluates plans against 15 risk criteria, quantifying violations like invalidated handoffs or proximity hazards. This approach not only mitigates risks but also provides interpretable safety reports, enabling continuous system improvement.

On the execution side in robotics, the motion and control policy must uphold the safety paradigm defined by the task planner. This safety paradigm arises from both the physical requirements of the robot—such as joint position, velocity, and torque limits—and the safety considerations associated with the operational and task spaces. This means that the control policy executing the task must prioritize safety over task performance, invoking additional safety protocols only when necessary to address actual safety violations. 

In our work, we integrate a Control Barrier Functions (CBFs) based control framework with a SAFER task planner to enforce safety within the control policy of robotic systems. Our motivation for utilizing CBFs stems from their strong theoretical guarantees \cite{ames2019control}. To the best of our knowledge, this is the first work that incorporates safety into an LLM-based task planner while integrating CBFs to ensure safety at the level of the robotic control policy. 

\textbf{Contributions. }We validate our framework through extensive experiments, including simulation tests on a heterogeneous robot task planning benchmark and real-world evaluations using two robot arms. Our experimental results demonstrate the practical effectiveness and enhanced safety of our approach. The key contributions of this paper are as follows:
\begin{itemize}
\item[\textbf{C1}] \emph{\proj Framework. }We propose a novel framework by incorporating a multi-LLM collaboration in the Planning Module. The framework ensures that safety checks are seamlessly embedded throughout the planning process, enabling more robust and safety-aware task execution.

\item[\textbf{C2}] \emph{Comprehensive Empirical Analysis. }Our approach is evaluated against non-\proj baselines using multiple reasoning models, both open-source and closed-source. Furthermore, we provide more observations and analysis on the cost and latency of these models in long-horizon robotics tasks.

\item[\textbf{C3}] \emph{Control Barrier Framework Integration. }We incorporate Control Barrier Functions (CBFs) within the \proj framework to enforce a safety-compliant control policy. This integration bridges the gap between high-level task planning and low-level control, ensuring that safety constraints are maintained during execution, even in dynamic environments.

\end{itemize}


%% file: Sections/Related_Works.tex
\section{Related Works}
\label{sec:Related}

\textbf{LLM based Task Planning in Robotics. }
Traditional task planning frameworks in robotics often struggled with real-world adaptability due to their limited reasoning capabilities~\cite{liu2024coherent}. Recently, LLMs have emerged as a promising alternative, demonstrating strong generalization in diverse settings such as household manipulation~\cite{ahn2022can, lin2023text2motion, liang2023code, driess2023palm} and navigation~\cite{yu2023co, long2024discuss}. Their ability to execute zero-shot~\cite{kojima2022large, huang2022language} and few-shot~\cite{brown2020language} reasoning further enhances their applicability. Additionally, LLMs have been leveraged for multi-robot collaboration, breaking down high-level commands into executable steps~\cite{shalarge, wang2024large}.

However, integrating LLMs into robotic task planning raises safety concerns, as they may generate contextually appropriate but hazardous plans. Recent work~\cite{li2024safe} has attempted to mitigate this by introducing safety-aware LLM planning pipelines. However, a major challenge in long-horizon tasks is the context-length limitation, which forces a trade-off between retaining task history for consistency and preserving critical safety constraints in prompts. Relying on a single LLM makes this trade-off unavoidable. To address this, we introduce a multi-LLM collaborative framework where different LLMs specialize in task planning and safety feedback. This approach ensures robust safety integration while maintaining long-term coherence, overcoming the limitations of single-LLM methods.

Safety has been a key focus in prior work related to LLM in robotics, such as \cite{yang2024plug, wang2024ensuring, ni2024don}. For instance, \cite{yang2024plug} employs Linear Temporal Logic (LTL) to enforce safety constraints, while \cite{wang2024ensuring} ensures safety at the level of motion policy. Similarly, \cite{ni2024don} addresses safety by accounting for various hazardous scenarios. However, these approaches fall short in addressing dynamically evolving environments or complex human-robot interactions, particularly when multiple humans with distinct roles impose varying safety requirements for others. To overcome these limitations, we propose a novel approach that embeds safety at the control level. This method ensures robust safety enforcement even in the presence of dynamic uncertainty and environmental changes.

The most well-known method for ensuring safety in robotics is Artificial Potential Fields (APF) \cite{khatib1986real}. APF employs an attraction-repulsion field to enforce safety, but this approach can persistently affect the control output, even when safety constraints are not active. Other notable works include \cite{fox1997dynamic}, which introduces velocity commands to achieve dynamic safety, and \cite{lacevic2013safety}, which proposes danger fields to quantify safety risks. However, these methods similarly exert continuous influence on the controller, regardless of whether safety constraints are relevant. In contrast, Control Barrier Functions (CBFs) have emerged as a promising alternative, offering theoretical safety guarantees \cite{ames2019control} and a demonstrated connection to APF \cite{singletary2021comparative}. CBFs ensure safety by minimally modifying the nominal controller, thereby limiting their impact when safety constraints are inactive. Moreover, CBFs enable real-time safety enforcement and have been successfully applied across various domains including bipedal \cite{hsu2015control}, manipulators \cite{murtaza2022safety, murtaza2021real}, swarm robotics \cite{wang2017safety} and quadrotors \cite{wang2017safe}.

\textbf{LLM-as-a-Judge. }LLM-as-a-Judge has emerged as a novel paradigm for evaluating the safety and feasibility of robotic task plans by leveraging the reasoning capabilities of LLMs~\cite{yin2024safeagentbench}. Unlike conventional rule-based safety checks, which rely on predefined constraints, LLM-as-a-Judge introduces a flexible and context-aware evaluation mechanism that can dynamically assess task plans against a set of safety principles. Prior works have explored LLM-based evaluation for programming correctness and factual consistency, but their application in robotic safety assessment is relatively new~\cite{tong2024codejudge}. By employing LLMs to quantify violations in generated task sequences, this approach provides a structured yet interpretable safety metric. Notably, this method enables real-time risk assessment by evaluating plans against multiple risk criteria, such as action dependencies, spatial constraints, and human-robot interaction risks. 
The integration of LLM-as-a-Judge within multi-LLM frameworks, as seen in \proj, represents a promising step toward robust safety-aware task planning, balancing efficiency with proactive error mitigation.


%% file: Sections/Framework.tex
\section{\proj: Safety-Aware Framework for Execution in Robotics}
\label{sec:Method}

\proj introduces a multi-LLM collaboration to enhance the safety and robustness of LLM-driven task planning. The framework is composed of four core modules: the Planning Module, Execution Module, and LLM-as-a-Judge, Feedback Module. These components work in collaboration to ensure that safety checks are systematically enforced throughout the task planning and execution pipeline.

\begin{figure}
    \centering
    \includegraphics[width=0.95\linewidth]{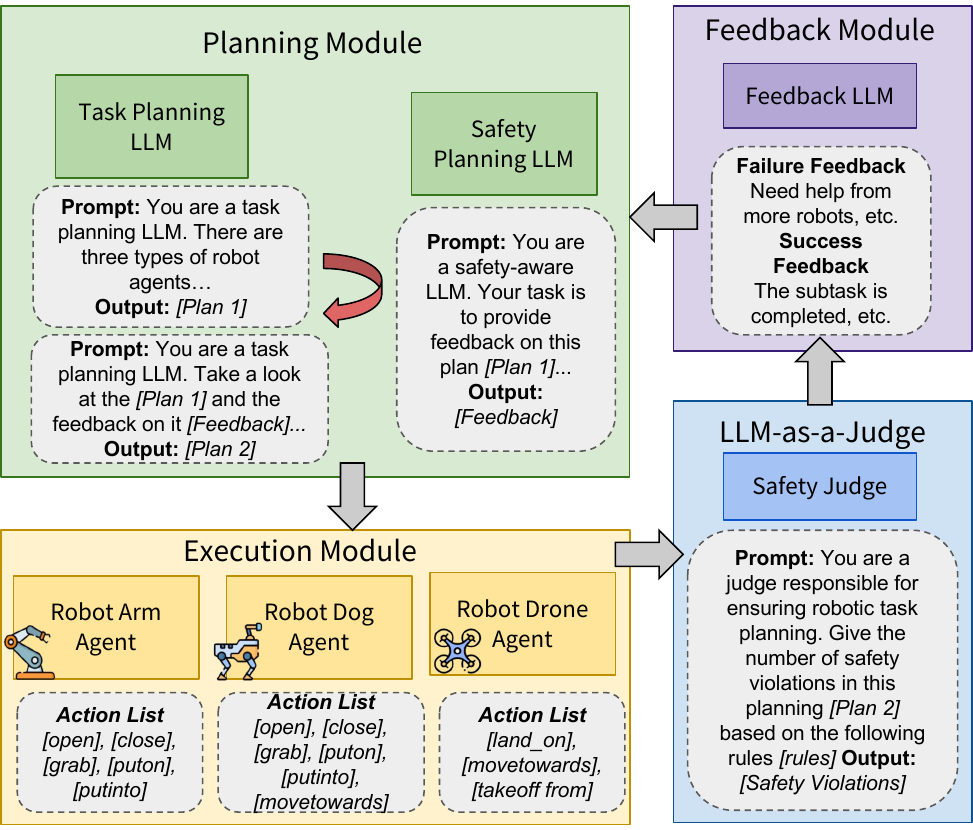}
    \caption{\textbf{Overview of \proj (Safety-Aware Framework for Execution in Robotics)}. The framework consists of four core modules: (1) Planning Module, where the Task Planning LLM generates initial task sequences while the Safety Planning LLM audits them for potential hazards and iteratively refines the plan; (2) Execution Module, responsible for deploying the task plan through heterogeneous robotic agents, each equipped with an Execution LLM that verifies execution; (3) LLM-as-a-Judge, a dedicated evaluation module that assesses the final task plan for safety violations, quantifying risk factors such as spatial conflicts and incomplete action sequences; and (3) Feedback Module, processes execution outcomes and provides corrective feedback to the Planning Module, ensuring real-time adaptation.}
    \label{fig:framework}
\end{figure}

\textbf{Planning Module. }As shown in Fig.~\ref{fig:framework} the Planning Module is responsible for generating structured safety-aware action sequences for heterogeneous robotic agents. Unlike single-agent planning models, \proj integrates two specialized LLMs to balance task efficiency with safety enforcement. It utilizes a Task Planning LLM that iteratively refines task plans based on external feedback.
The Task Planning LLM serves as the primary orchestrator for action sequencing. Given a high-level task description, agent capabilities, and environmental observations, the Task Planning LLM decomposes the goal into executable subtasks. To introduce safety awareness, the Safety Planning LLM works in collaboration with the primary planner. This safety-focused LLM provides safety-related feedback on the generated task sequence to identify potential hazards, including spatial conflicts, invalid action dependencies, and omitted preconditions. The feedback generated by this LLM is integrated  into the planning LLM, ensuring that the final plan adheres to generalized safety constraints while remaining efficient.

\textbf{(RQ)} \emph{Why multi-LLM collaboration over Retrieval-Augmented Generation (RAG) or Few-Shot Learning (FSL)?} While techniques like RAG and FSL can help retrieve relevant information and integrate it into the prompt of a central task planner, they do not address the context window limitation inherent in LLMs. These approaches still rely on fitting all necessary details within a single LLM's prompt, which constrains long-horizon reasoning and safety enforcement. In contrast, our multi-LLM collaboration approach decouples task planning and safety feedback, allowing one LLM to focus on structured task generation while another LLM specializes in safety assessment and feedback.

\textbf{Execution Module. }The Execution Module is responsible for deploying the refined task plans into actionable commands for robotic agents. It consists of heterogeneous robot agents (robot arms, quadrotor, and robotic dog) that execute the planned actions. Each robot agent is equipped with one specific Robot Execution LLM. As shown in Fig.~\ref{fig:framework}, the robot agents have a list of actions they can perform. Instead of conventional execution pipelines that rigidly follow predefined trajectories, \proj enables real-time adaptability by verifying the feasibility of actions before execution.
Once tasks are assigned to specific robot agents, the Robot Execution LLM predicts whether the task is executable. If a task fails, the Feedback Module analyzes the failure and sends the feedback to the planning module.

\textbf{Feedback Module. }As shown in Figure~\ref{fig:framework}, the feedback module takes the current state input including task goals, robot capabilities, and execution progress. The module suggests corrections after action failures and progress updates after successful executions. The module generated two forms of feedback: \textit{Failure Feedback} and  \textit{Success Feedback}. The \textit{failure feedback} is triggered when an action cannot be executed due to environmental constraints, insufficient resources, or incorrect sequencing. Examples include a robotic arm failing to grasp an object due to misalignment or a drone unable to take off due to an obstructed flight path. Then in the next step, the planning module refines the task plan by utilizing the output from the feedback module.
The \textit{success feedback} is issued when a subtask is successfully completed, allowing the transition to the next execution phase. This feedback ensures that \proj maintains coherence across long-horizon tasks.

%% file: Sections/controlbarrierfunction.tex
\section{Safety and Control}
\label{sec:CBF}
To ensure compliance with the safety metrics prescribed by \proj, we employ Control Barrier Functions (CBFs). CBFs ensure safety by first designing a safe set, then using the CBFs to ensure forward invariance of the given set i.e. if the system starts with the safe set, it remains in the safe set. CBFs provide theoretical guarantees regarding safety and ensure safety by modifying the nominal controller in a minimally invasive manner. Its architecture can be applied to several control strategies and fast solvers ensure real-time deployment. This ensures safety even when the reference trajectory violates safety constraints. We have designed the constraints for the robotics system following approaches similar to those in \cite{murtaza2022safety, murtaza2021real}. Here, we briefly review the CBFs and associated safety constraints referenced from SAFER for safe task planning. For a comprehensive mathematical understanding, readers are directed to the original papers in \cite{ames2019control, murtaza2022safety, murtaza2021real}.

\subsection{Control Barrier Functions}
For most robotics systems, the safety constraints of the robotics systems can be defined in terms of their state. States space associated with robotics can be defined in terms of its joint, task, or operational space. As CBFs require first defining a safe set, hence the safe set needs to be designed in terms of the associated joint, task, or operational space. If $x$ is the state associated with joint, task, or operational space, and $\mathscr{C}_0$ defines the safe set such that barrier function, $h(x) \geq 0$, ensures safety. Then the safe set can be written as
\begin{equation*}\label{Eq:SafeSet}
    \mathscr{C}_0 = \{x\in \mathcal{T}~ \vert~ h(x) \ge 0\}
\end{equation*}
where $\mathcal{T}$ represents the associated joint, task, or operational space. For most of the robotics systems, input appears by taking one or two derivatives of the safe set. For the second-order systems, the equation is given as 
\begin{equation*}
\ddot{h}(x) = L_f^2h(x) + L_fL_gh(x)u
\end{equation*}
where $L_f^2h(x)$ and $L_fL_gh(x)$ are the Lie derivatives of $h(x)$ and $u$ is the input associated with robotics system. The safety associated with the barrier function, $h(x)$, can then be ensured if the following inequality is ensured
\begin{equation*}
    L_f^2h(x) + L_fL_gh(x)u + K\eta \geq 0
\end{equation*}
where $\eta = \left[h(x) \dot{h}(x)\right]$, and $\dot{h}(x)$ is the first derivative of the barrier function $h(x)$. The coefficients of the matrix, $K$, are chosen by the principles of pole placement in a closed loop linear system and are defined in more detail in \cite{ames2019control}. For the first-order constraint, the inequality simplifies to 
\begin{equation*}
    L_fh(x) + L_gh(x)u + \gamma h(x) \geq 0
\end{equation*}
where $\gamma$ is a positive number. If we have a nominal controller, $u^{nom}$, we can ensure the safety constraint by minimally modifying the nominal controller by solving a Quadratic Programming (QP) based optimization problem to yield a safe controller as 
\begin{equation}\label{Eq:QP-CBF}
\begin{aligned}
&~~~~~~~~ u^{*} =  \underset{u}{\argmin}
~~\frac{1}{2}\Vert u - u^{nom}\Vert^{2} && \\
& ~~~~~~~\text{subject to}:\\
&~~~~~~~~~~~~~ L_{f}^{2}h(x) + L_gL_fh(x)u + K\eta  \ge 0 &&\\
\end{aligned}
\end{equation}
The flexibility of the QP-based optimization is that it allows additional constraints associated with the robotics system as well. The safety constraints can be broadly classified into two categories, joint and operational space. We will next define the safety constraints associated with the robotics system without delving into the mathematics of each constraint. 

\subsection{Joint Safety Constraints}
The safety constraints associated with the hardware properties of the robotics systems fall into the category of joint safety constraints. These safety constraints include joint position limits, joint velocity limits, and torque limits associated with motors for each robotics system. They are generally defined by the hardware manufacturer, and any violation of these constraints results in an inoperative state of the robotics system. 

\subsection{Task and Operational Safety Constraints}
Task and Operational safety constraints associated with robotics systems include properties like obstacle avoidance, limiting the operational space of the robots as well as singularity avoidance associated with robotics systems such that the robots do not lose their ability to control in any direction. It also includes constraints like not colliding with other robots and objects. These constraints ensure the safety of the controller, the safety of the human involved, and the safety of the robots from being damaged. 

\subsection{Goals and Constraints Setting}
The LLMs define a sequence of instructions and a set of constraints for each robot. At each step on the sequence, we introduce a parser that translates the robot's subtask into a goal, either defining the desired velocity of the base, pose of the end-effector, or the gripper's state, among others. These goals are updated to achieve each of the steps on the sequence. For the set of constraints, they might be global, which are active through the whole sequence, or might be activated at a given step. The constraints are also transformed using a parser that defines barrier functions $h(x)$ that the optimizer will have to consider during control.

%% file: Sections/Experiments.tex
\section{Experiments and Results}
\label{sec:Experiments}
We evaluate \proj using the COHERENT benchmark for multi-agent robotic task planning~\cite{liu2024coherent}. This benchmark encompasses complex long-horizon tasks that require collaboration between heterogeneous robotic agents, making it an ideal testbed for assessing the effectiveness of safety-aware planning.

\textbf{Benchmark. }The COHERENT benchmark consists of various task scenarios requiring multi-agent coordination. Tasks are categorized according to the number of robots required to complete them as mono-type, dual-type, and trio-type tasks. Mono-type tasks require a single robot, dual-type tasks require two robots, and trio-type tasks require three robots to complete the task. 
Similar to our Non-Safety baseline \cite{liu2024coherent}, we use 40 tasks considering the high cost of closed-source model APIs calls. From each scene, we pick 8 tasks with 2 mono-type, 3 dual-type, and 3 trio-type tasks.

\textbf{Evaluation Metrics. }To quantitatively evaluate \proj's performance, we use steps and safety violations. Steps highlight the number of execution steps required to complete a task, reflecting efficiency. Safety Violation measures of the number of detected safety violations per task. 
A task refers to the overall goal that needs to be achieved, whereas steps represent the individual actions or subtasks that must be executed sequentially to accomplish the task.
In Table~\ref{tab:main_table}, we report the average of the two metrics over a scene. These metrics provide a comprehensive evaluation framework that captures both task execution efficiency and safety awareness.

\input{Tables/main_table_v2}

\textbf{Results and Discussion. } Table~\ref{tab:main_table} presents a comparative evaluation of \proj against a non-safety baseline across the COHERENT benchmark, using both GPT-4o and DeepSeek-r1 models. The results indicate that \proj significantly reduces safety violations in all scenes while maintaining competitive task efficiency. Specifically, for \proj using the DeepSeek-r1 model (S-4o) achieves a reduction of 77.5\% in ASV, demonstrating its effectiveness in enforcing safety rules. Similarly, for the GPT-4o model, \proj reduces ASV by 47\% across all scenarios, underscoring its generalization across different LLMs as planners.

\begin{figure}[ht]
  \centering
      \includegraphics[width=0.65\linewidth]{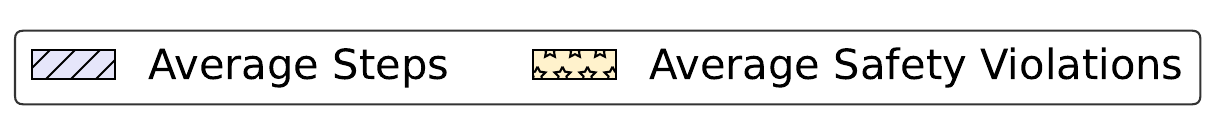}
  \begin{subfigure}[t]{0.4\textwidth}
    \centering
    \includegraphics[width=\linewidth]{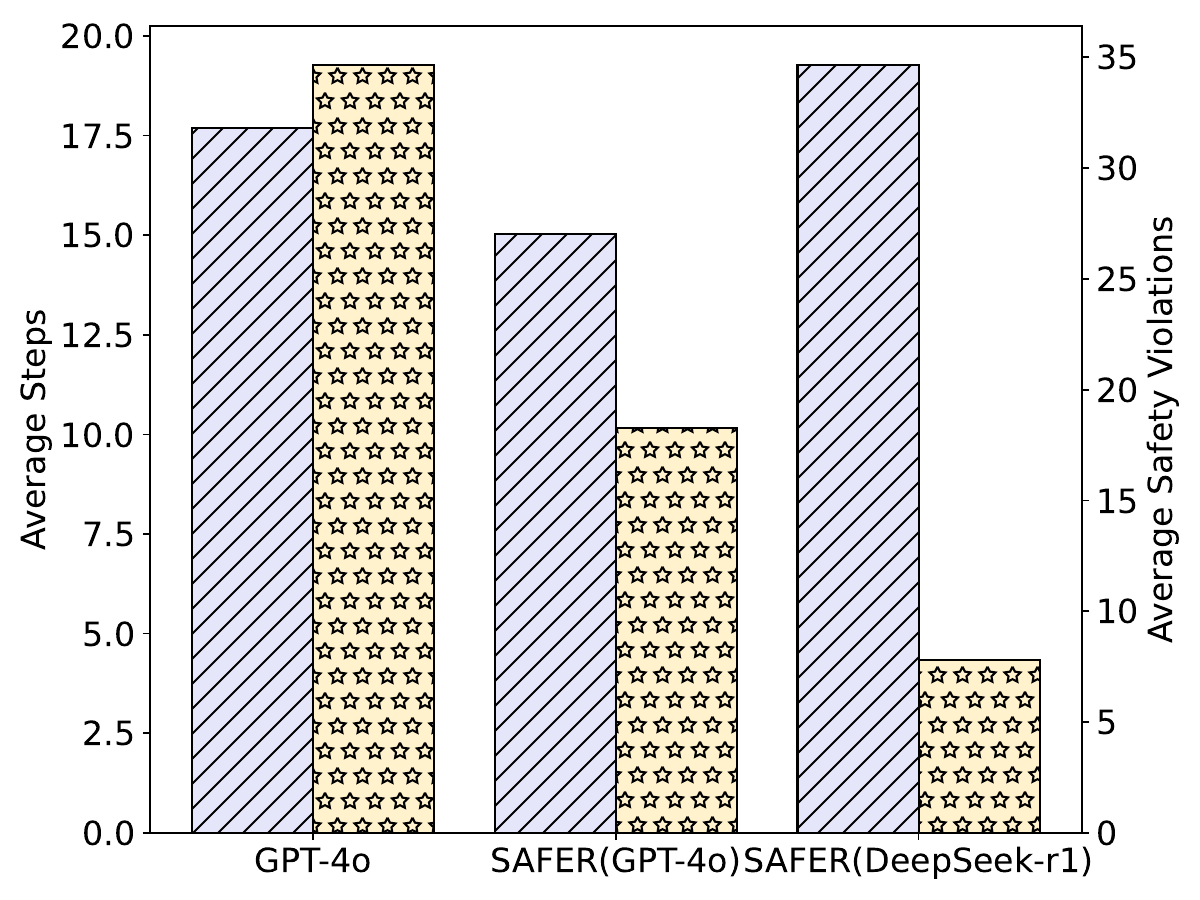}
  \end{subfigure}%
  \caption{\textbf{Comparative evaluation of \proj vs. Non-safety baseline.} The figure shows the Average Steps and Average Safety Violations across task scenarios (S1–S5) for baseline and \proj, highlighting improved safety with minimal impact on efficiency.}
  \label{fig:plot}
\end{figure}



\begin{figure}[ht]
  \centering
      \includegraphics[width=0.65\linewidth]{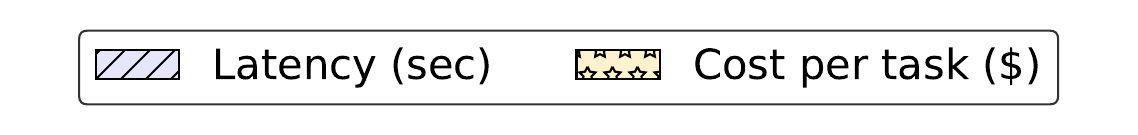}
  \begin{subfigure}[t]{0.24\textwidth}
    \centering
    \includegraphics[width=\linewidth]{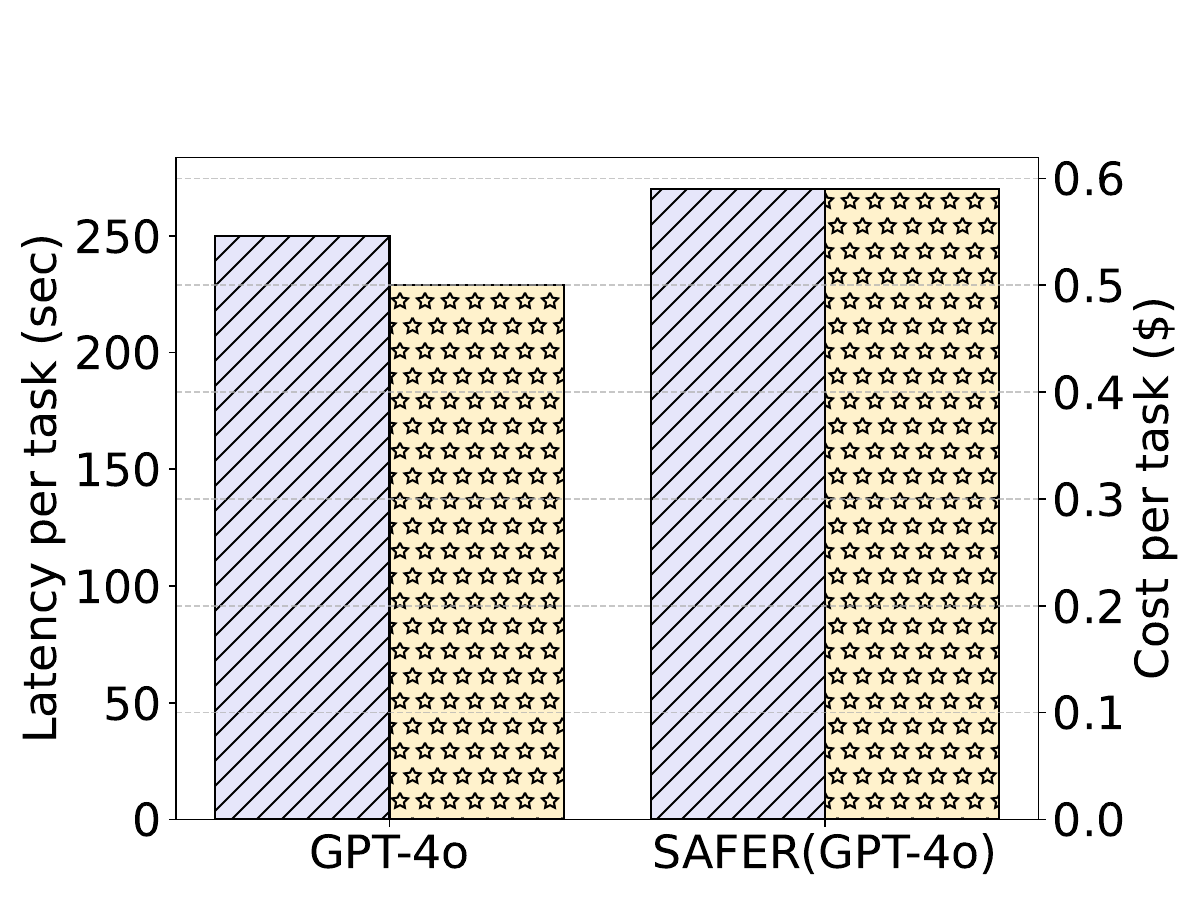}
  \end{subfigure}%
  \hfill
  \begin{subfigure}[t]{0.24\textwidth}
    \centering
    \includegraphics[width=\linewidth]{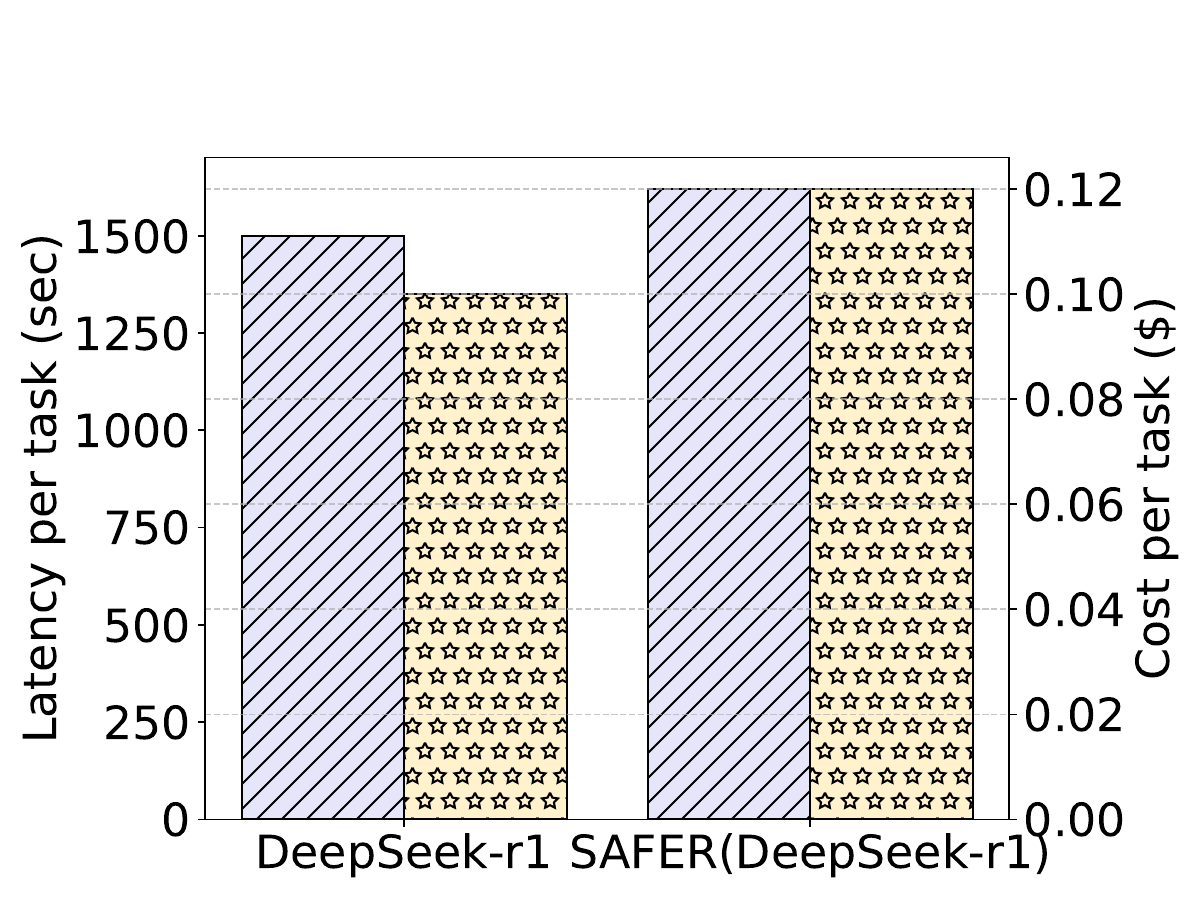}
  \end{subfigure}%
  \caption{Latency and cost comparison of \proj with baseline models. The left plot compares \proj with GPT-4o, while the right plot compares \proj with DeepSeek-r1, illustrating low overheads introduced by the safety-aware framework.}
  \label{fig:latency}
  \vspace{-1em}
\end{figure}


Figure~\ref{fig:plot} highlights the trade-off between task efficiency and safety enforcement in different models. The figure shows the average results across all the scenes in the benchmark. The figure demonstrates a clear reduction in safety violations when using \proj, validating its ability to mitigate execution risks. However, the non-safety baseline (GPT-4o) exhibits the highest ASV. Moreover, the impact of \proj's remains consistent between different LLM models, reinforcing its adaptability. This reinforces the central claim that a multi-LLM safety-aware framework can enhance robotic task execution without incurring excessive efficiency trade-offs. 
Despite better reasoning \proj(DeepSeek-r1) required more steps compared to \proj(GPT-4o). This is because its improved reasoning significantly reduces safety violations, leading the model to smaller steps to task completion. 
Furthermore, our results reveal some interesting observations. 

Figure~\ref{fig:latency} highlights the latency and cost overheads introduced by \proj for GPT-4o and DeepSeek-r1. Although \proj enhances safety enforcement, it incurs only a minimal increase in computational time and cost. The magnitude of this overhead remains consistent across models. DeepSeek-r1 experiences higher latency due to its sensitivity to load. Furthermore, the consistent overhead is attributed to \proj introducing two API calls per step.

\textbf{OB\ding{182} Stronger reasoning abilities translate to better safety-aware task planning abilities.} We observed that DeepSeek-r1, which has stronger reasoning abilities than GPT-4o, also exhibits significantly fewer safety violations. The reason for this is that more capable reasoners likely generate more structured and foresighted plans, anticipating safety risks in advance rather than reacting to them later because of longer chains that they can generate. 

\textbf{OB\ding{183} LLMs-as-a-Safety-Judge are fair judges rather than answer validators.} Previous research has shown that LLMs, when used as answer judges, tend to favor their own responses~\cite{panickssery2025llm, xu2024pride}. However, we do not see this trend when LLMs are used as safety judges. The safety assessment may involve a more explicit rule-based check, where objectivity is required. This shift in task nature may reduce the bias seen in answer validation settings.

\textbf{OB\ding{184} Balancing efficiency and safety is achievable with a multi-LLM framework.} Our results show that \proj effectively reduces safety violations without introducing excessive steps to complete the tasks. While enforcing safety constraints often leads to longer execution times, the better reasoning capabilities of multi-LLM frameworks help mitigate this trade-off. By leveraging multiple models, \proj can make informed adjustments that optimize both safety and task efficiency, rather than sacrificing one for the other. This suggests that carefully designed multi-LLM interactions can enable safer task execution without compromising overall performance.

\textbf{OB\ding{185} LLMs Tend to Prioritize Fast and Efficient Plans Over Detailed Safety Checks.} Our experiments show that when LLMs are not given clear instructions to consider safety, they naturally generate plans that are quick and efficient. This means that while the generated plans may require fewer steps, they often miss important safety details. In simpler terms, if safety guidelines aren't explicitly provided, LLMs focus on getting the job done fast rather than carefully checking for potential hazards. This finding highlights the importance of incorporating explicit safety directives when using LLMs in environments where risk management is crucial.


\section{Real Robot Deployment}
\label{sec:Real_Robot}

\begin{figure}
    \centering
    \includegraphics[width=0.95\linewidth]{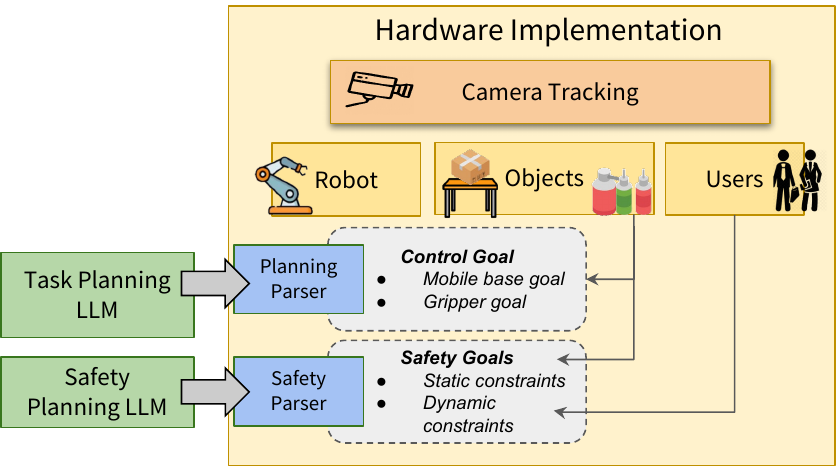}
    \caption{Overview of the hardware implementation.}
    \label{fig:hardware_implementation}
\end{figure}

\begin{figure}
    \centering
    \includegraphics[width=0.95\linewidth]{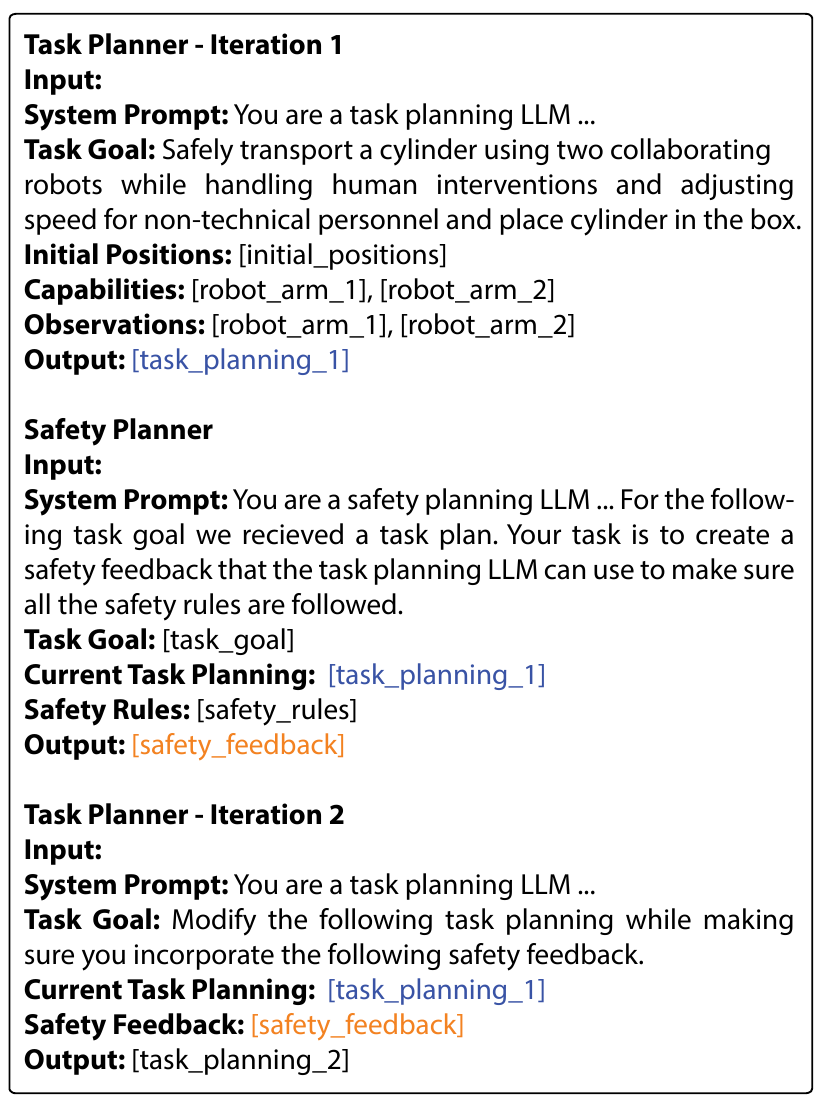}
    \caption{Prompts and Output representation of Multi-LLM Collaboration in the Planning Module.}
    \label{fig:prompt_rep}
\end{figure}

The outputs generated by LLMs, such as GPT-4o, are expressed in natural language. A representation of the information contained in the prompts is shown in Fig.~\ref{fig:prompt_rep}. 
To convert these outputs into actionable instructions for robotic we propose parsing the natural language descriptions into structured commands, mapping them to robotic control APIs, and ensuring seamless execution. Our approach focuses on creating a robust pipeline that bridges the gap between high-level task specifications and low-level robotic actions while maintaining high accuracy. To accomplish this, we introduce two parsers that consider the robot capabilities, along with objects and users' pose to define the control and safety goals as shown in Fig.~\ref{fig:hardware_implementation}.

\paragraph{Parsing Task Planning LLM Outputs}
This LLM outputs a sequence of instructions, such as:
\texttt{Move Robot 1 to table A, Robot 1 pick the can, Move Robot 1 to table B, Robot 2 pick the box, Robot 2 move box to center of table, Robot 1 place can in box.}
These outputs are parsed into structured commands using a custom parser. We separate instructions into their respective robots, identify objects and describe them by their pose in the world, and grasp configuration, finally transform actions into goals. For example, \texttt{Move Robot 1 to table A}, table A has position $(0.0, 1.2, 1.0)$ and the action \texttt{Move} translate into a new goal pose for the base in $SE(2)$ as $(0.0, 0.6, 90^\circ)$. Other instructions are further split into a subsequence of steps, e.g. \texttt{Robot 1 pick the can}, the can has a position $(0.05, 1.1, 1.1)$ and a preconfigured grasp for the gripper to grasp the object. The \texttt{pick} instruction is divided into $4$ steps: align of gripper over the object, reach for object, close gripper, and lift object. 

\paragraph{Parsing Safety Planning LLM Outputs}
This LLM output a set of constraints for individual robots, such as:
\texttt{Robot 1 manipulator must stay away from users, Robot 1 must not collide with table A, Robot 2 must not collide with table B}. These outputs are parsed into structured definitions of control barrier constraints. We can define both dynamic and static constraints. E.g. \texttt{not collide with Table A} requires a single query of the table A's position and dimension of the table and we can create a static polygon that encapsulates the object and create a hard constraint for the robot's motion. While \texttt{stay away from user} is a dynamic constraint that updates as a user moves and can have different dimensions depending on the user's access level. The robot is constantly checking the user's relative position and the controller ensures that the arm will not violate the user's safe space.



\paragraph{Robot Experiment}
\begin{figure*}
    \centering
    \includegraphics[width=0.99\linewidth]{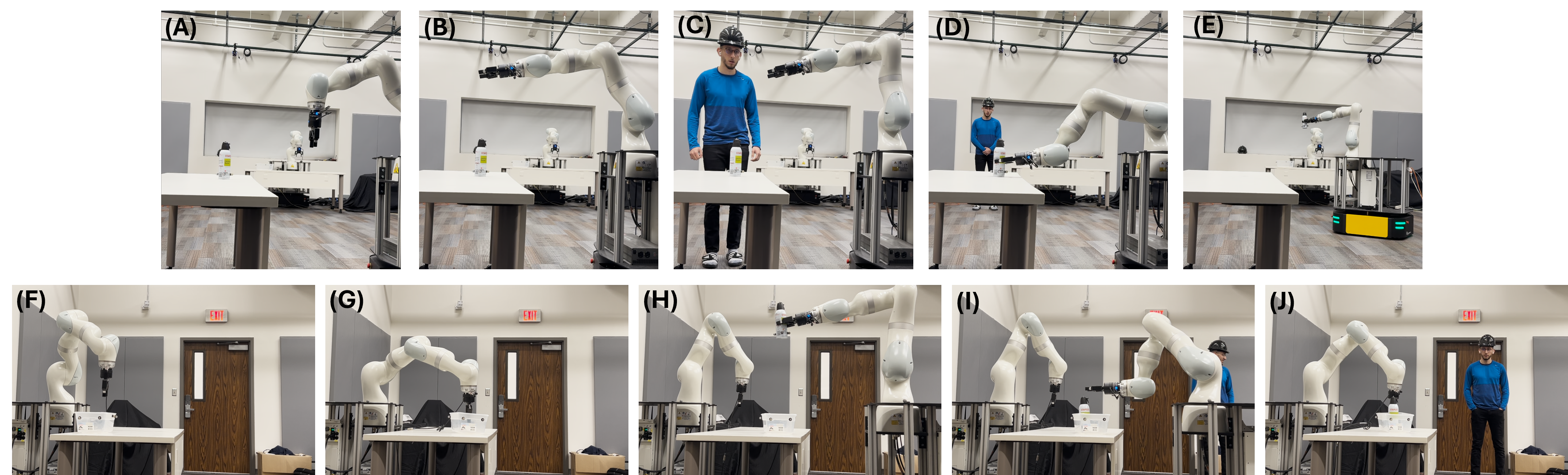}
    \caption{Steps of the hardware experiment. (A) Robot 1 approaches table A. (B) Robot 1 aligns gripper with spray can. (C) As a person approaches the Robot 1, the manipulator moves back. (D) Robot 1 picks the spray can. (E) Robot 1 moves to table B. (F) Robot 2 aligns gripper with box. (G) Robot 2 picks the box and brings it to the center of the table. (H) Robot 2 leaves box and moves to safe position, while Robot 1 gets closer. (I) Robot 1 places spray can in the box. (J) Robot 1 leaves to the initial position and Robot 2 moves the box to a new location.}
    \label{fig:sequence}
\end{figure*}
For our experimental setup, we are using two mobile manipulators that consider a Kuka IIWA LBR 14 robotic arms, which consist of 7 DoF and a clearpath Ridgeback omnidirectional mobile base. Each arm has a two-fingers gripper from Robotiq. For object and robot tracking, we are using a Vicon camera tracking system. For the static objects as tables, we have their global position, for the interacting objects like the spray can and box, we have place markers to track them. To track users, we are using a tracking helmet that gives us the overall position of the person in the workspace. The mobile base uses a PD controller to command the desired twist for the base, wrapped with a CBF. The robotic arm is using the proposed force/torque controller described in \ref{sec:CBF}.

For the hardware experiment, we consider two robots, one that can move about the environment, while the second is placed at a fixed point. There are two types of users in the environment that should be considered: $i)$ non-technical personnel, which the robot must keep a distance from, and $ii)$ technical personnel that the robot must slow down its motion when close. We consider the following task: Safely transport a cylinder using two collaborating robots while handling human interventions and adjusting speed for non-technical personnel, and place the cylinder in the box. 

The summarized planned task is as follows:
Robot Arm 1 navigates to the cylinder using obstacle avoidance, ensures safety, and aligns precisely before lowering and securely grabbing it. It then lifts the cylinder, transports it to the Robot Arm 2 workspace, checks for clearance. Robot Arm 2 verifies safety, aligns, and picks up the box and places it at the center of Table B. Robot Arm 2 releases the box and moves to a safe location. Robot Arm 1, upon arrival at Table B, it checks Robot Arm 2 to be at a safe position, then aligns, lowers, and releases the cylinder safely into the box. Robot Arm 1 moves away from the workspace. Robot Arm 2 verifies that Robot Arm 1 is at a safe position, then aligns, lowers, and picks up the box and places it close to the user. If needed, Robot Arm 1 moves the cylinder cautiously, keeping a distance from non-technical personnel.

We present the hardware execution of the resulting sequence by the task planning LLM in Fig.~\ref{fig:sequence} and in the supporting video. Along the sequence, the system is modifying the goal for the mobile base to move the Robot 1 around the space and setting the end-effector's goals depending on the task. Along with the basic task allocation, the safety LLM also activates different control barriers that will help maintain safety standards. Over the whole experiment, the robot is able to progress through the sequence, achieving each task and subtasks and successfully finishing the requested task.

%% file: Tables/main_table_v2.tex
\renewcommand{\arraystretch}{1.2}
\definecolor{pink}{HTML}{EAD5DC}
\definecolor{blue}{HTML}{EEE3E7}
\definecolor{lightgreen}{HTML}{E8F0E6}
\definecolor{lightred}{HTML}{F4E8E6}


\renewcommand{\arraystretch}{1.2} 
\setlength{\tabcolsep}{2.05pt} 

\begin{table}[t]
    \centering
    \footnotesize 
    \begin{tabular*}{\linewidth}{l|cl|cl|cl|cl|cl}
        \hline
        \rowcolor{pink} & \multicolumn{2}{c|}{\textbf{S1}} 
        & \multicolumn{2}{c|}{\textbf{S2}} 
        & \multicolumn{2}{c|}{\textbf{S3}} 
        & \multicolumn{2}{c|}{\textbf{S4}} 
        & \multicolumn{2}{c}{\textbf{S5}} \\
        \rowcolor{pink} \textbf{Model} & \textbf{AS} & \textbf{ASV} & \textbf{AS} & \textbf{ASV} & \textbf{AS} & \textbf{ASV} & \textbf{AS} & \textbf{ASV} & \textbf{AS} & \textbf{ASV} \\
        \hline
        \textbf{WS-4o} & 15.8 & 32.6 & 12.1 & 23.0  & 22.3 & 51.5  & 12.3 & 26.9  & 26.0 & 39.3  \\
        \textbf{S-4o}   & 12.1 & 17.6 \textcolor{green}{$\downarrow$}  & 11.9 & 17.1 \textcolor{green}{$\downarrow$}  & 14.5 & 16.8 \textcolor{green}{$\downarrow$} & 14.6 & 20.4 \textcolor{green}{$\downarrow$} & 22.0 & 19.5 \textcolor{green}{$\downarrow$}  \\
        \textbf{S-r1}   & 20.5 & \,\,\,5.8 \textcolor{green}{$\downarrow$}  
                   & 15.4 & \,\,\,6.4 \textcolor{green}{$\downarrow$}  
                   & 22.6 & 12.8 \textcolor{green}{$\downarrow$}  
                   & 19.3 & \,\,\,5.9 \textcolor{green}{$\downarrow$}  
                   & 18.6 & \,\,\,8.3 \textcolor{green}{$\downarrow$}  \\
        \hline
    \end{tabular*}
    \caption{\textbf{Comparison of \proj on COHERENT Benchmark against non-safety baseline for GPT-4o and DeepSeek-r1 models}. Abbreviations: \textbf{S1-S5} (Scenes from COHERENT benchmark), \textbf{AS} (Average Steps), \textbf{ASV} (Average Safety Violations), \textbf{WS-4o} (GPT-4o model without \proj), \textbf{S-4o} (\proj using GPT-4o model), \textbf{S-r1} (\proj using DeepSeek-r1 model). Arrows indicate change from WS-4o: \textcolor{green}{$\downarrow$} indicates improvements.}
    \label{tab:main_table}
    \vspace{-1em} 
\end{table}

%% file: Sections/DiscussionConclusion.tex
\section{Conclusions}
\label{sec:conclusion}
In this work, we introduced \proj designed to enhance safety in LLM-driven task planning. By employing a multi-LLM collaboration approach, \proj integrates a dedicated Safety Planning LLM and an LLM-as-a-Judge module to systematically assess and mitigate risks throughout the planning and execution pipeline. Our framework ensures that safety checks are embedded into task plans while maintaining task efficiency. We evaluated \proj against non-safety baselines in complex multi-robot scenarios, demonstrating a significant reduction in safety violations with minimal impact on execution efficiency. Additionally, we integrate CBFs to enforce safety constraints at the low-level control stage, ensuring real-time safety enforcement. Experimental results, including both simulations and real-world deployments, validate the effectiveness of our approach in reducing hazardous failures while maintaining adaptability across different robotic agents. Future work will explore extending \proj to more dynamic environments, incorporating adaptive safety feedback, and refining LLM-based evaluation to further enhance transparency and reliability in autonomous robotic systems.